\documentclass[11pt]{article}

\usepackage[preprint]{acl}

\usepackage{times}
\usepackage{latexsym}

\usepackage[T1]{fontenc}

\usepackage[utf8]{inputenc}

\usepackage{microtype}

\usepackage{inconsolata}

\usepackage{graphicx}
\usepackage{amsmath}
\usepackage{makecell}
\usepackage{enumitem}
\usepackage[table,xcdraw]{xcolor}

\usepackage[utf8]{inputenc} 
\usepackage[T1]{fontenc}    
\usepackage{hyperref}       
\usepackage{url}            
\usepackage{booktabs}       
\usepackage{amsfonts}       
\usepackage{nicefrac}       
\usepackage{microtype}      
\usepackage{xcolor}         
\usepackage{booktabs}
\usepackage[table]{xcolor}
\usepackage[normalem]{ulem}
\usepackage{multirow}
\usepackage{algorithm}
\usepackage{algpseudocode}
\algtext*{EndFor}

\usepackage{listings}

\usepackage{color}

\lstdefinestyle{concalprompt}{
    basicstyle=\small\ttfamily,
    breaklines=true,
    breakatwhitespace=true,
    columns=fullflexible,
    keepspaces=true,
    showstringspaces=false,
    frame=single,
    framerule=0.4pt,
    literate={‑}{{-}}1
}

\title{Action-Value Calibration via Continuation Sharing for Long-Horizon Agentic Reinforcement Learning}

\author{
 \textbf{Yingjian Chen\textsuperscript{1,2}},
 \textbf{Lang Zhou\textsuperscript{1,2}},
 \textbf{Panbo He}\textsuperscript{2},
 \textbf{Zixin Zhao\textsuperscript{1,2}},
 \\
 \textbf{Pingfan Li\textsuperscript{1,2}},
 \textbf{Weishi Zheng\textsuperscript{1,2}},
 \textbf{Zhilin Zhao\textsuperscript{1,2}}\thanks{Corresponding author}
\\
\\
 \textsuperscript{1}Sun Yat-sen University,
 \textsuperscript{2}Shenzhen Loop Area Institute
}

\newcommand{\paperrevision}{0}
\begin{document}
\def\sectionsuffix{}
\ifnum\paperrevision=0\relax
\else
  \ifnum\paperrevision=2\relax
    \def\sectionsuffix{_v2}
  \else
    \PackageError{concal}{Unsupported paper revision \paperrevision}{Set \string\paperrevision\space to 0 or 2.}
  \fi
\fi
\maketitle
\input{sections/abstract\sectionsuffix}

\input{sections/introduction\sectionsuffix}
\input{sections/related_work\sectionsuffix}
\input{sections/preliminaries\sectionsuffix}
\input{sections/method\sectionsuffix}
\input{sections/experiments\sectionsuffix}
\input{sections/conclusion\sectionsuffix}
\input{sections/limitations\sectionsuffix}
\section*{Use of Generative AI Tools}
Generative AI tools were used for language refinement and to assist with figure preparation and coding. The authors reviewed the resulting content and take full responsibility for the methodology, results, and conclusions.

\bibliography{custom}

\begin{thebibliography}{34}
\providecommand{\natexlab}[1]{#1}

\bibitem[{Achiam et~al.(2023)Achiam, Adler, Agarwal, Ahmad, Akkaya, Aleman, Almeida, Altenschmidt, Altman, Anadkat et~al.}]{openai2023gpt4}
Josh Achiam, Steven Adler, Sandhini Agarwal, Lama Ahmad, Ilge Akkaya, Florencia~Leoni Aleman, Diogo Almeida, Janko Altenschmidt, Sam Altman, Shyamal Anadkat, et~al. 2023.
\newblock {GPT-4} technical report.
\newblock \emph{arXiv preprint arXiv:2303.08774}.

\bibitem[{Ahmadian et~al.(2024)Ahmadian, Cremer, Gall{\'e}, Fadaee, Kreutzer, Pietquin, {\"U}st{\"u}n, and Hooker}]{ahmadian-etal-2024-back}
Arash Ahmadian, Chris Cremer, Matthias Gall{\'e}, Marzieh Fadaee, Julia Kreutzer, Olivier Pietquin, Ahmet {\"U}st{\"u}n, and Sara Hooker. 2024.
\newblock Back to basics: Revisiting {REINFORCE}-style optimization for learning from human feedback in {LLMs}.
\newblock In \emph{Proceedings of the 62nd Annual Meeting of the Association for Computational Linguistics (Volume 1: Long Papers)}, pages 12248--12267.

\bibitem[{Bai et~al.(2024)Bai, Zhou, Cemri, Pan, Suhr, Levine, and Kumar}]{bai2024digirl}
Hao Bai, Yifei Zhou, Mert Cemri, Jiayi Pan, Alane Suhr, Sergey Levine, and Aviral Kumar. 2024.
\newblock {DigiRL}: Training in-the-wild device-control agents with autonomous reinforcement learning.
\newblock In \emph{Advances in Neural Information Processing Systems}, volume~37, pages 12461--12495.

\bibitem[{Bai et~al.(2025)Bai, Chen, Liu, Wang, Ge, Song, Dang, Wang, Wang, Tang, Zhong, Zhu, Yang, Li, Wan, Wang, Ding, Fu, Xu, Ye, Zhang, Xie, Cheng, Zhang, Yang, Xu, and Lin}]{bai2025qwen25vl}
Shuai Bai, Keqin Chen, Xuejing Liu, Jialin Wang, Wenbin Ge, Sibo Song, Kai Dang, Peng Wang, Shijie Wang, Jun Tang, Humen Zhong, Yuanzhi Zhu, Mingkun Yang, Zhaohai Li, Jianqiang Wan, Pengfei Wang, Wei Ding, Zheren Fu, Yiheng Xu, and 8 others. 2025.
\newblock Qwen2.5-vl technical report.

\bibitem[{Carta et~al.(2023)Carta, Romac, Wolf, Lamprier, Sigaud, and Oudeyer}]{carta2023grounding}
Thomas Carta, Cl{\'e}ment Romac, Thomas Wolf, Sylvain Lamprier, Olivier Sigaud, and Pierre-Yves Oudeyer. 2023.
\newblock Grounding large language models in interactive environments with online reinforcement learning.
\newblock In \emph{International Conference on Machine Learning}, pages 3676--3713.

\bibitem[{Cheng et~al.(2026)Cheng, He, Feng, Xu, Yan, Feng, and An}]{cheng2026graphgpo}
Xin Cheng, Shuo He, Lang Feng, HaiYang Xu, Ming Yan, Lei Feng, and Bo~An. 2026.
\newblock Beyond trajectory-level attribution: Graph-based credit assignment for agentic reinforcement learning.
\newblock \emph{arXiv preprint arXiv:2605.26684}.

\bibitem[{Comanici et~al.(2025)Comanici, Bieber, Schaekermann, Pasupat, Sachdeva, Dhillon, Blistein, Ram, Zhang, Rosen et~al.}]{comanici2025gemini}
Gheorghe Comanici, Eric Bieber, Mike Schaekermann, Ice Pasupat, Noveen Sachdeva, Inderjit Dhillon, Marcel Blistein, Ori Ram, Dan Zhang, Evan Rosen, et~al. 2025.
\newblock Gemini 2.5: Pushing the frontier with advanced reasoning, multimodality, long context, and next generation agentic capabilities.
\newblock \emph{arXiv preprint arXiv:2507.06261}.

\bibitem[{Feng et~al.(2025)Feng, Xue, Liu, and An}]{feng2025gigpo}
Lang Feng, Zhenghai Xue, Tingcong Liu, and Bo~An. 2025.
\newblock Group-in-group policy optimization for {LLM} agent training.
\newblock In \emph{Advances in Neural Information Processing Systems}, volume~38, pages 46375--46408.

\bibitem[{He et~al.(2024)He, Yao, Ma, Yu, Dai, Zhang, Lan, and Yu}]{he-etal-2024-webvoyager}
Hongliang He, Wenlin Yao, Kaixin Ma, Wenhao Yu, Yong Dai, Hongming Zhang, Zhenzhong Lan, and Dong Yu. 2024.
\newblock {WebVoyager}: Building an end-to-end web agent with large multimodal models.
\newblock In \emph{Proceedings of the 62nd Annual Meeting of the Association for Computational Linguistics (Volume 1: Long Papers)}, pages 6864--6890.

\bibitem[{He et~al.(2026)He, Feng, Wei, Cheng, Feng, and An}]{he2026hgpo}
Shuo He, Lang Feng, Qi~Wei, Xin Cheng, Lei Feng, and Bo~An. 2026.
\newblock Hierarchy-of-groups policy optimization for long-horizon agentic tasks.
\newblock In \emph{International Conference on Learning Representations}, volume 2026, pages 27572--27593.

\bibitem[{Hurst et~al.(2024)Hurst, Lerer, Goucher, Perelman, Ramesh, Clark, Ostrow, Welihinda, Hayes, Radford et~al.}]{openai2024gpt4o}
Aaron Hurst, Adam Lerer, Adam~P Goucher, Adam Perelman, Aditya Ramesh, Aidan Clark, AJ~Ostrow, Akila Welihinda, Alan Hayes, Alec Radford, et~al. 2024.
\newblock {GPT-4o} system card.
\newblock \emph{arXiv preprint arXiv:2410.21276}.

\bibitem[{Ichter et~al.(2023)Ichter, Brohan, Chebotar, Finn, Hausman, Herzog, Ho, Ibarz, Irpan, Jang, Julian, Kalashnikov, Levine, Lu, Parada, Rao, Sermanet, Toshev, Vanhoucke, Xia, Xiao, Xu, Yan, Brown, Ahn, Cortes, Sievers, Tan, Xu, Reyes, Rettinghouse, Quiambao, Pastor, Luu, Lee, Kuang, Jesmonth, Joshi, Jeffrey, Ruano, Hsu, Gopalakrishnan, David, Zeng, and Fu}]{ichter2023saycan}
Brian Ichter, Anthony Brohan, Yevgen Chebotar, Chelsea Finn, Karol Hausman, Alexander Herzog, Daniel Ho, Julian Ibarz, Alex Irpan, Eric Jang, Ryan Julian, Dmitry Kalashnikov, Sergey Levine, Yao Lu, Carolina Parada, Kanishka Rao, Pierre Sermanet, Alexander~T Toshev, Vincent Vanhoucke, and 26 others. 2023.
\newblock Do as {I} can, not as {I} say: Grounding language in robotic affordances.
\newblock In \emph{Conference on Robot Learning}, pages 287--318.

\bibitem[{Ji et~al.(2026)Ji, Ma, Wang, Chen, Chu, and Wu}]{ji2026tree}
Yuxiang Ji, Ziyu Ma, Yong Wang, Guanhua Chen, Xiangxiang Chu, and Liaoni Wu. 2026.
\newblock Tree search for {LLM} agent reinforcement learning.
\newblock In \emph{International Conference on Learning Representations}, volume 2026, pages 87362--87388.

\bibitem[{Jin et~al.(2025)Jin, Zeng, Yue, Yoon, Arik, Wang, Zamani, and Han}]{jin2025searchr1}
Bowen Jin, Hansi Zeng, Zhenrui Yue, Jinsung Yoon, Sercan Arik, Dong Wang, Hamed Zamani, and Jiawei Han. 2025.
\newblock {Search-R1}: Training {LLMs} to reason and leverage search engines with reinforcement learning.
\newblock \emph{arXiv preprint arXiv:2503.09516}.

\bibitem[{Li et~al.(2026)Li, Wang, Yan, Tian, Xu, Song, Xu, and Cheong}]{li-etal-2026-salt}
Jiazheng Li, Yawei Wang, Qiaojing Yan, Yijun Tian, Zhichao Xu, Huan Song, Panpan Xu, and Lin~Lee Cheong. 2026.
\newblock {SALT}: Step-level advantage assignment for long-horizon agents via trajectory graph.
\newblock In \emph{Findings of the Association for Computational Linguistics: EACL 2026}, pages 4709--4725.

\bibitem[{Liu et~al.(2025)Liu, Chen, Li, Qi, Pang, Du, Lee, and Lin}]{liu2025understanding}
Zichen Liu, Changyu Chen, Wenjun Li, Penghui Qi, Tianyu Pang, Chao Du, Wee~Sun Lee, and Min Lin. 2025.
\newblock Understanding {R1-Zero}-like training: A critical perspective.
\newblock \emph{arXiv preprint arXiv:2503.20783}.

\bibitem[{Ouyang et~al.(2022)Ouyang, Wu, Jiang, Almeida, Wainwright, Mishkin, Zhang, Agarwal, Slama, Ray, Schulman, Hilton, Kelton, Miller, Simens, Askell, Welinder, Christiano, Leike, and Lowe}]{ouyang2022training}
Long Ouyang, Jeffrey Wu, Xu~Jiang, Diogo Almeida, Carroll Wainwright, Pamela Mishkin, Chong Zhang, Sandhini Agarwal, Katarina Slama, Alex Ray, John Schulman, Jacob Hilton, Fraser Kelton, Luke Miller, Maddie Simens, Amanda Askell, Peter Welinder, Paul~F Christiano, Jan Leike, and Ryan Lowe. 2022.
\newblock Training language models to follow instructions with human feedback.
\newblock In \emph{Advances in Neural Information Processing Systems}, volume~35, pages 27730--27744.

\bibitem[{{Qwen} et~al.(2025){Qwen}, Yang, Yang, Zhang, Hui, Zheng, Yu, Li, Liu, Huang, Wei, Lin, Yang, Tu, Zhang, Yang, Yang, Zhou, Lin, Dang, Lu, Bao, Yang, Yu, Li, Xue, Zhang, Zhu, Men, Lin, Li, Tang, Xia, Ren, Ren, Fan, Su, Zhang, Wan, Liu, Cui, Zhang, and Qiu}]{qwen2.5}
{Qwen}, An~Yang, Baosong Yang, Beichen Zhang, Binyuan Hui, Bo~Zheng, Bowen Yu, Chengyuan Li, Dayiheng Liu, Fei Huang, Haoran Wei, Huan Lin, Jian Yang, Jianhong Tu, Jianwei Zhang, Jianxin Yang, Jiaxi Yang, Jingren Zhou, Junyang Lin, and 24 others. 2025.
\newblock {Qwen2.5} technical report.
\newblock \emph{arXiv preprint arXiv:2412.15115}.

\bibitem[{Schick et~al.(2023)Schick, Dwivedi-Yu, Dess{\`\i}, Raileanu, Lomeli, Hambro, Zettlemoyer, Cancedda, and Scialom}]{schick2023toolformer}
Timo Schick, Jane Dwivedi-Yu, Roberto Dess{\`\i}, Roberta Raileanu, Maria Lomeli, Eric Hambro, Luke Zettlemoyer, Nicola Cancedda, and Thomas Scialom. 2023.
\newblock {Toolformer}: Language models can teach themselves to use tools.
\newblock In \emph{Advances in Neural Information Processing Systems}, volume~36, pages 68539--68551.

\bibitem[{Schrader(2018)}]{schrader2018gymsokoban}
Max-Philipp~B. Schrader. 2018.
\newblock {gym-sokoban}.
\newblock \url{https://github.com/mpSchrader/gym-sokoban}.

\bibitem[{Schulman et~al.(2017)Schulman, Wolski, Dhariwal, Radford, and Klimov}]{schulman2017ppo}
John Schulman, Filip Wolski, Prafulla Dhariwal, Alec Radford, and Oleg Klimov. 2017.
\newblock Proximal policy optimization algorithms.
\newblock \emph{arXiv preprint arXiv:1707.06347}.

\bibitem[{Shao et~al.(2024)Shao, Wang, Zhu, Xu, Song, Bi, Zhang, Zhang, Li, Wu, and Guo}]{shao2024deepseekmath}
Zhihong Shao, Peiyi Wang, Qihao Zhu, Runxin Xu, Junxiao Song, Xiao Bi, Haowei Zhang, Mingchuan Zhang, Y.~K. Li, Y.~Wu, and Daya Guo. 2024.
\newblock {DeepSeekMath}: Pushing the limits of mathematical reasoning in open language models.
\newblock \emph{arXiv preprint arXiv:2402.03300}.

\bibitem[{Shinn et~al.(2023)Shinn, Cassano, Gopinath, Narasimhan, and Yao}]{shinn2023reflexion}
Noah Shinn, Federico Cassano, Ashwin Gopinath, Karthik Narasimhan, and Shunyu Yao. 2023.
\newblock {Reflexion}: Language agents with verbal reinforcement learning.
\newblock In \emph{Advances in Neural Information Processing Systems}, volume~36, pages 8634--8652.

\bibitem[{Shridhar et~al.(2021)Shridhar, Yuan, C{\^o}t{\'e}, Bisk, Trischler, and Hausknecht}]{shridhar2021alfworld}
Mohit Shridhar, Xingdi Yuan, Marc-Alexandre C{\^o}t{\'e}, Yonatan Bisk, Adam Trischler, and Matthew Hausknecht. 2021.
\newblock {ALFWorld}: Aligning text and embodied environments for interactive learning.
\newblock In \emph{International Conference on Learning Representations}.

\bibitem[{Wang et~al.(2025)Wang, Wang, Wang, Zhang, Li, Yang, Jin, Yu, Nguyen, Liu, Gottlieb, Lu, Cho, Wu, Fei-Fei, Wang, Choi, and Li}]{wang2025ragen}
Zihan Wang, Kangrui Wang, Qineng Wang, Pingyue Zhang, Linjie Li, Zhengyuan Yang, Xing Jin, Kefan Yu, Minh~Nhat Nguyen, Licheng Liu, Eli Gottlieb, Yiping Lu, Kyunghyun Cho, Jiajun Wu, Li~Fei-Fei, Lijuan Wang, Yejin Choi, and Manling Li. 2025.
\newblock {RAGEN}: Understanding self-evolution in {LLM} agents via multi-turn reinforcement learning.
\newblock \emph{arXiv preprint arXiv:2504.20073}.

\bibitem[{Wei et~al.(2025)Wei, Yao, Liu, Zhang, Lu, Qiu, Yu, Xu, Zhang, Yin, Yun, and Li}]{wei-etal-2025-webagent}
Zhepei Wei, Wenlin Yao, Yao Liu, Weizhi Zhang, Qin Lu, Liang Qiu, Changlong Yu, Puyang Xu, Chao Zhang, Bing Yin, Hyokun Yun, and Lihong Li. 2025.
\newblock {WebAgent-R1}: Training web agents via end-to-end multi-turn reinforcement learning.
\newblock In \emph{Proceedings of the 2025 Conference on Empirical Methods in Natural Language Processing}, pages 7909--7928.

\bibitem[{Xi et~al.(2025)Xi, Ding, Chen, Hong, Guo, Wang, Guo, Yang, Liao, He, Gao, Chen, Zheng, Zou, Gui, Zhang, Qiu, Huang, Wu, and Jiang}]{xi-etal-2025-agentgym}
Zhiheng Xi, Yiwen Ding, Wenxiang Chen, Boyang Hong, Honglin Guo, Junzhe Wang, Xin Guo, Dingwen Yang, Chenyang Liao, Wei He, Songyang Gao, Lu~Chen, Rui Zheng, Yicheng Zou, Tao Gui, Qi~Zhang, Xipeng Qiu, Xuanjing Huang, Zuxuan Wu, and Yu-Gang Jiang. 2025.
\newblock {AgentGym}: Evaluating and training large language model-based agents across diverse environments.
\newblock In \emph{Proceedings of the 63rd Annual Meeting of the Association for Computational Linguistics (Volume 1: Long Papers)}, pages 27914--27961.

\bibitem[{Xi et~al.(2026)Xi, Huang, Liao, Huang, Liu, Guo, Yang, Zheng, Ye, Zhang, Chen, He, Ding, Li, Chen, Du, Yao, Xu, Chen, Gui, Wu, Zhang, Huang, and Jiang}]{xi2026agentgymrl}
Zhiheng Xi, Jixuan Huang, Chenyang Liao, Baodai Huang, Jiaqi Liu, Honglin Guo, Yajie Yang, Rui Zheng, Junjie Ye, Jiazheng Zhang, Wenxiang Chen, Wei He, Yiwen Ding, Guanyu Li, Zehui Chen, Zhengyin Du, Xuesong Yao, Yufei Xu, Jiecao Chen, and 5 others. 2026.
\newblock {AgentGym-RL}: An open-source framework to train {LLM} agents for long-horizon decision making via multi-turn {RL}.
\newblock In \emph{International Conference on Learning Representations}.

\bibitem[{Yao et~al.(2022)Yao, Chen, Yang, and Narasimhan}]{yao2022webshop}
Shunyu Yao, Howard Chen, John Yang, and Karthik Narasimhan. 2022.
\newblock {WebShop}: Towards scalable real-world web interaction with grounded language agents.
\newblock In \emph{Advances in Neural Information Processing Systems}, volume~35, pages 20744--20757.

\bibitem[{Yao et~al.(2023)Yao, Zhao, Yu, Du, Shafran, Narasimhan, and Cao}]{yao2023react}
Shunyu Yao, Jeffrey Zhao, Dian Yu, Nan Du, Izhak Shafran, Karthik Narasimhan, and Yuan Cao. 2023.
\newblock {ReAct}: Synergizing reasoning and acting in language models.
\newblock In \emph{International Conference on Learning Representations}.

\bibitem[{Yao et~al.(2026)Yao, Fu, Liu, and Zhang}]{yao2026graft}
Xincheng Yao, Haobo Fu, Weiming Liu, and Chongyang Zhang. 2026.
\newblock Back to the definition: Estimating step-level advantages via trajectory graphs for agentic reinforcement learning.
\newblock \emph{arXiv preprint arXiv:2609.28963}.

\bibitem[{Yu et~al.(2025)Yu, Zhang, Zhu, Yuan, Zuo, Yue, Dai, Fan, Liu, liu, Liu, Liu, Lin, Lin, Ma, Sheng, Tong, Zhang, Zhang, Zhang, Zhang, Zhu, Zhu, Chen, Chen, Wang, Yu, Song, Wei, Zhou, Liu, Ma, Zhang, Yan, Wu, and Wang}]{yu2025dapo}
Qiying Yu, Zheng Zhang, Ruofei Zhu, Yufeng Yuan, Xiaochen Zuo, Yu~Yue, Weinan Dai, Tiantian Fan, Gaohong Liu, juncai liu, LingJun Liu, Xin Liu, Haibin Lin, Zhiqi Lin, Bole Ma, Guangming Sheng, Yuxuan Tong, Chi Zhang, Mofan Zhang, and 17 others. 2025.
\newblock {DAPO}: An open-source {LLM} reinforcement learning system at scale.
\newblock In \emph{Advances in Neural Information Processing Systems}, volume~38, pages 113222--113244.

\bibitem[{Zheng et~al.(2026)Zheng, Sun, Bao, Liu, Jiang, Song, and Dou}]{zheng2026pgpo}
Yuyao Zheng, Haipeng Sun, Junwei Bao, Lemao Liu, Hongfei Jiang, Yang Song, and Dejing Dou. 2026.
\newblock {PGPO}: Potential-guided policy optimization for multi-turn agentic tasks.
\newblock \emph{arXiv preprint arXiv:2609.02236}.

\bibitem[{Zhou et~al.(2024)Zhou, Zanette, Pan, Levine, and Kumar}]{zhou2024archer}
Yifei Zhou, Andrea Zanette, Jiayi Pan, Sergey Levine, and Aviral Kumar. 2024.
\newblock {ArCHer}: Training language model agents via hierarchical multi-turn {RL}.
\newblock In \emph{Proceedings of the 41st International Conference on Machine Learning}, volume 235, pages 62178--62209.

\end{thebibliography}

\clearpage
\appendix

\section*{Appendix}
\input{sections/appendix\sectionsuffix}
\end{document}